\pdfoutput=1
\documentclass[11pt]{article}

\usepackage[margin=1in]{geometry}
\usepackage[T1]{fontenc}
\usepackage[utf8]{inputenc}
\usepackage{lmodern}
\usepackage{microtype}
\usepackage{amsmath,amssymb,amsthm,mathtools}
\usepackage{booktabs}
\usepackage{hyperref}
\usepackage{xcolor}
\usepackage{graphicx}
\usepackage{setspace}

\hypersetup{
    colorlinks=true,
    linkcolor=blue,
    citecolor=blue,
    urlcolor=blue
}

\newtheorem{definition}{Definition}
\newtheorem{proposition}{Proposition}
\newtheorem{remark}{Remark}

\title{A Public Theory of Distillation Resistance via Constraint-Coupled Reasoning Architectures}
\author{%
Peng WEI \\College of Plant Protection \\Southwest University, China \\\texttt{weipeng2019@swu.edu.cn}
\and
Wesley Shu \\The Institute of Energetic Paradigm \\\texttt{shuwesley@gmail.com}}

\date{March 2026}

\begin{document}
\maketitle

\begin{abstract}
Knowledge distillation, model extraction, and behavior transfer have become central concerns in frontier AI. The main risk is not merely copying, but the possibility that useful capability can be transferred more cheaply than the governance structure that originally accompanied it. This paper presents a public, trade-secret-safe theoretical framework for reducing that asymmetry at the architectural level. The core claim is that distillation becomes less valuable as a shortcut when high-level capability is coupled to internal stability constraints that shape state transitions over time. To formalize this idea, the paper introduces a constraint-coupled reasoning framework with four elements: bounded transition burden, path-load accumulation, dynamically evolving feasible regions, and a capability--stability coupling condition. The paper is intentionally public-safe: it omits proprietary implementation details, training recipes, thresholds, hidden-state instrumentation, deployment procedures, and confidential system design choices. The contribution is therefore theoretical rather than operational. It offers a falsifiable architectural thesis, a clear threat model, and a set of experimentally testable hypotheses for future work on distillation resistance, alignment, and model governance.
\end{abstract}


\section{Introduction}

Knowledge distillation was originally proposed as a method for transferring predictive structure from a large teacher model to a smaller student model while preserving much of the teacher's practical performance \cite{hinton2015}. Over time, distillation expanded from a deployment-efficiency technique into a broader paradigm of capability transfer. In large language models, this now includes the transfer of task competence, instruction-following behavior, reasoning traces, and other useful response patterns \cite{gou2021,hsieh2023}. That shift has strategic implications. Once high-value behavior can be reproduced at lower cost, the economic and governance barriers around advanced capability become more fragile.

The central problem is not simply imitation. The deeper issue is an asymmetry between capability and governance. A frontier model is usually developed under substantial cost: training compute, safety work, evaluation, refusal design, monitoring assumptions, and deployment controls. Yet many of the useful behavioral outputs of that model may still be compressible. If a smaller or differently governed system can recover a substantial portion of those capabilities through distillation or extraction, then the benefits of the original governance regime do not travel with equal fidelity. In that case, capability is more portable than control.

Existing AI safety practice addresses this problem mainly through external or semi-external measures. These include reinforcement learning from human feedback, constitutional objectives, instruction tuning, system-level filtering, API access control, logging, and provenance or watermarking mechanisms \cite{christiano2017,ouyang2022,bai2022,kirchenbauer2023}. These measures are important. However, most do not directly answer a more structural question: can the internal route by which a model becomes highly capable be designed so that useful capability does not compress cleanly when the stabilizing structure is omitted?

This paper develops a public theoretical answer to that question. The argument is that capability transfer becomes harder, or at least less attractive, when strong performance depends on internal stability constraints that shape the sequence of latent transitions rather than only the observable output. In such a design, useful behavior is not merely a surface pattern; it is the visible trace of a constrained internal process. Distillation then faces a more difficult task. It must preserve not only answers, but the underlying route geometry that made those answers stable.

The paper does not claim that such a system has already been fully implemented or empirically validated at frontier scale. Nor does it disclose confidential technical mechanisms. Instead, it offers a trade-secret-safe public formulation: a theoretical framework, a threat model, a mathematical scaffold, and a set of falsifiable hypotheses. That is the correct level of claim for a public arXiv version.

\section{Why Distillation Is a Governance Problem}

Distillation is often discussed as an optimization problem. A student model is trained to reproduce the behavior of a teacher under parameter, latency, or data constraints. From a governance perspective, however, the key issue is different. The question is whether a student can recover useful capability without proportionally recovering the stability and control structure that originally constrained it.

Let $T$ denote a teacher model and $S$ denote a student model. Standard distillation seeks parameters $\phi$ for $S$ such that
\begin{equation}
\mathbb{E}_{x \sim \mathcal{D}} \left[d\!\left(S_{\phi}(x), T(x)\right)\right] \leq \varepsilon,
\label{eq:distill}
\end{equation}
where $d(\cdot,\cdot)$ is a task-relevant discrepancy and $\mathcal{D}$ is an input distribution. This formulation captures behavioral similarity. It does not by itself capture whether the conditions that made the teacher reliable, robust, or governable are also reproduced.

To express that gap, let $\mathcal{K}(M)$ denote the useful capability profile of model $M$, and let $\mathcal{G}(M)$ denote its governance-relevant stability profile. The latter may include robustness to prompt perturbation, bounded amplification, refusal consistency, internal stability under longer reasoning trajectories, or other safety-relevant properties. In practice, one often faces a pattern of the following form:
\begin{equation}
\Delta\!\left(\mathcal{K}(S),\mathcal{K}(T)\right) \text{ is small,}
\qquad
\Delta\!\left(\mathcal{G}(S),\mathcal{G}(T)\right) \text{ is not small.}
\label{eq:asymmetry}
\end{equation}
This is the capability--governance asymmetry. It is the structural concern around distillation.

When a model is safe mainly because of external wrappers or output-level policy shaping, the useful core may remain highly compressible. A student can then inherit competence more easily than governance. This is why the problem should not be framed only as one of copying. It is a problem of uneven transferability.

\section{Related Work}

The present framework connects to several lines of literature.

First, knowledge distillation has a long history as a model compression technique \cite{hinton2015,gou2021}. Subsequent work showed that a student can sometimes match or approach teacher performance even under significant compression, and that teacher guidance can improve generalization beyond straightforward supervised learning \cite{furlanello2018}.

Second, the alignment literature studies how to shape model behavior toward human preferences or safety constraints. Important examples include preference-based reinforcement learning, instruction tuning, and Constitutional AI \cite{christiano2017,ouyang2022,bai2022}. These methods improve output behavior, but they do not automatically guarantee that the model's useful competence is inseparable from the stabilizing structure imposed during training or deployment.

Third, model extraction and stealing attacks show that useful functional behavior can often be recovered from black-box or gray-box access \cite{tramer2016,orekondy2019,zanella2021}. This reinforces the concern that high-value capability may diffuse more easily than its original governance setting.

Fourth, reasoning-oriented distillation and trace-based training show that structured intermediate signals can be highly informative for capability transfer \cite{wei2022,hsieh2023}. That result is useful for performance, but it also raises the strategic value of internal process information.

Finally, watermarking and provenance methods seek to detect or deter unauthorized reuse \cite{kirchenbauer2023}. These methods are valuable, but they do not by themselves alter the internal compressibility of the capability being transferred.

The present paper differs from all of these by focusing on architecture-level resistance. The aim is not only to regulate outputs, detect reuse, or limit access. The aim is to make useful capability less separable from the internal structure that stabilizes it.

\section{Threat Model and Scope}

This paper considers a public threat model rather than a classified or proprietary one. The relevant adversary is an actor attempting to recover useful capability from a teacher model or model family through behavior imitation, output harvesting, structured trace collection, or distillation-like compression. The goal of the adversary is not necessarily bit-level replication. It is enough to recover a large fraction of the useful performance profile under lower cost or weaker governance.

The framework does not claim protection against every attack class. It does not claim immunity against direct weight theft, insider compromise, or unrestricted white-box replication. Nor does it claim that architecture alone can replace institutional safety, evaluation, or legal protection. The narrower claim is that under a meaningful class of compression and behavior-transfer attacks, the value of distillation should decline if useful capability is strongly coupled to internal stability constraints.

This scope matters for scientific precision. A public arXiv paper should not overclaim. The correct statement is not that extraction becomes impossible in all settings. The statement is that the transfer frontier changes when stability is built into the route to competence itself.

\section{Constraint-Coupled Reasoning: Public Formulation}

The central idea of this paper is that a capable reasoning model should not be understood only in terms of output quality. It should also be understood as a system whose internal state transitions remain within stability-preserving constraints over time. In that case, useful performance is not just a function of endpoint behavior; it is a property of the path by which the system arrives there.

This section presents a public-safe mathematical abstraction. The formulation is intentionally high level. It omits proprietary definitions of hidden variables, internal probes, threshold schedules, and optimization routines.

\subsection{State evolution}

Let $h_t \in \mathbb{R}^n$ denote the latent state at step $t$, and let $x_t$ denote the current input token or control signal. The model evolves according to
\begin{equation}
h_{t+1} = F_{\theta}(h_t, x_t),
\label{eq:evolution}
\end{equation}
where $\theta$ denotes model parameters.

A standard view of reasoning focuses on the output induced by this recurrence. The present framework instead asks whether the transition sequence itself is regulated in a way that matters for stable competence.

\subsection{Transition burden}

Associate with each state transition a nonnegative burden quantity
\begin{equation}
b_t = \Psi(h_t, h_{t+1}, x_t),
\label{eq:burden}
\end{equation}
where $\Psi$ is a burden functional. In a concrete implementation, $\Psi$ could depend on activation concentration, routing sharpness, sensitivity, instability indicators, or other internal quantities. Those specifics are intentionally omitted here. For the public formulation, only the abstract role matters: $b_t$ measures how costly or destabilizing a transition is relative to the intended stability regime.

The first constraint is local boundedness:
\begin{equation}
b_t \leq B_t,
\label{eq:localbound}
\end{equation}
where $B_t$ is an admissible threshold schedule.

This means that useful reasoning is not allowed to rely on arbitrarily unstable or overloaded latent jumps.

\subsection{Path-load accumulation}

Define the cumulative path load
\begin{equation}
L_t = \sum_{\tau=1}^{t} \alpha_{\tau} b_{\tau},
\label{eq:pathload}
\end{equation}
where $\alpha_{\tau} > 0$ are weighting coefficients. The role of $L_t$ is to encode path dependence. A reasoning system is not determined only by its current state. It is also shaped by how much structured burden has accumulated along the path that led there.

This is important because many failures in reasoning systems are not purely local. A model may appear stable over short windows while gradually accumulating instability along a longer chain.

\subsection{Dynamic feasible region}

Let $\Omega_t \subseteq \mathbb{R}^n$ denote the feasible region of stable next states at time $t$. The public formulation assumes
\begin{equation}
\Omega_t = \Omega_0 \setminus \Gamma(L_t),
\label{eq:feasible}
\end{equation}
where $\Gamma(\cdot)$ is a contraction operator that removes portions of state space as accumulated path load grows.

The next transition is constrained by
\begin{equation}
h_{t+1} \in \Omega_t.
\label{eq:inomega}
\end{equation}

This means the model's future reasoning options are dynamically shaped by its previous burden profile. Stability is therefore endogenous to the state evolution rather than added only at the output layer.

\subsection{Capability--stability coupling}

The key public claim can now be expressed as a coupling condition. Let $\mathcal{R}(M)$ denote the stability profile of a model family, meaning the degree to which high performance depends on satisfying the burden and feasibility constraints above. A model family is said to satisfy capability--stability coupling if, for any teacher $T$ and student $S$ in the family,
\begin{equation}
\Delta\!\left(\mathcal{K}(S), \mathcal{K}(T)\right) \leq \varepsilon
\quad \Longrightarrow \quad
\Delta\!\left(\mathcal{R}(S), \mathcal{R}(T)\right) \leq g(\varepsilon),
\label{eq:coupling}
\end{equation}
where $g(\varepsilon) \to 0$ as $\varepsilon \to 0$.

In words, if a student gets close in useful capability, it must also get close in the internal stability structure that supports that capability. This is the opposite of the usual asymmetry in Equation~\eqref{eq:asymmetry}. The aim is not absolute impossibility of imitation. The aim is to reduce the extent to which high-value capability can be preserved while the stabilizing architecture is discarded.

\begin{definition}
A reasoning architecture is \emph{constraint-coupled} if useful capability depends materially on burden-bounded, path-dependent state evolution of the form given by Equations~\eqref{eq:localbound}--\eqref{eq:coupling}.
\end{definition}

\begin{definition}
A teacher architecture is \emph{distillation-resistant in the public sense} if, under a specified attack class, preserving capability within a small tolerance requires preserving a correspondingly close stability profile.
\end{definition}

These definitions are deliberately relative to the attack class and metric choice. That makes the claim experimentally meaningful rather than rhetorical.

\section{Why Constraint Coupling Should Change Distillation Dynamics}

Ordinary distillation succeeds when output behavior is a sufficiently informative summary of the teacher's useful competence. The student does not need to reproduce the teacher's full internal route; it only needs to learn a compressed function that matches the teacher well enough on relevant inputs.

Constraint-coupled reasoning changes that geometry. If strong performance depends on remaining within dynamically constrained regions of latent state space, then the observable output is no longer a complete summary of what matters. It becomes the surface expression of a regulated internal process. A student trained only on input-output traces may still imitate short-horizon behavior, but it faces greater difficulty in reproducing the deeper path-dependent conditions that support robustness.

This leads to the following mechanism claim.

\begin{proposition}
Suppose a teacher model $T$ derives a nontrivial portion of its useful performance from constraint-coupled state evolution. Suppose further that a student model $S$ is trained primarily to match observable teacher behavior without faithfully reconstructing the teacher's underlying stability structure. Then at least one of the following outcomes should occur:

\begin{itemize}
    \item the student retains a nontrivial capability gap,
    \item the student retains a nontrivial stability gap,
    \item the student requires disproportionate hidden burden to approach teacher-level performance.
\end{itemize}
\end{proposition}

\begin{remark}
This is not an impossibility theorem. It is a falsifiable mechanism hypothesis. The stronger the teacher's capability--stability coupling, the less likely it is that output imitation alone can cheaply preserve both capability and stability.
\end{remark}

The public significance of this proposition is straightforward. If it holds, then distillation becomes a less attractive shortcut for recovering useful but weakly governed capability. That is already valuable, even if perfect resistance is unattainable.

\section{A Public Training Objective Sketch}

A public paper should disclose the architecture thesis without revealing proprietary optimization details. The following objective therefore serves only as a schematic.

Let $\mathcal{L}_{\text{task}}$ denote the ordinary task or next-token loss. A generic constraint-coupled objective can be written as
\begin{equation}
\mathcal{L}_{\text{total}}
=
\mathcal{L}_{\text{task}}
+
\lambda_1 \sum_t \max\{0, b_t - B_t\}
+
\lambda_2 \sum_t \Phi(h_{t+1}, \Omega_t)
+
\lambda_3 \mathcal{L}_{\text{stab}},
\label{eq:objective}
\end{equation}
where:
\begin{equation}
b_t = \Psi(h_t, h_{t+1}, x_t),
\qquad
L_t = \sum_{\tau=1}^{t} \alpha_{\tau} b_{\tau},
\qquad
\Omega_t = \Omega_0 \setminus \Gamma(L_t).
\end{equation}

Here, the second term penalizes local burden violations, the third penalizes transitions that leave or approach the boundary of the feasible region, and the fourth captures additional stability objectives. The details of $\Psi$, $\Gamma$, $\Phi$, and $\mathcal{L}_{\text{stab}}$ are intentionally omitted. The purpose of the public formulation is to specify the structure of the claim, not to expose proprietary implementation.

\section{Public Experimental Program}

A public theory paper should make clear how its claims could be tested.

The simplest experimental route is to compare two teacher families trained on the same task distribution:
\begin{equation}
T_{\text{base}} \quad \text{and} \quad T_{\text{cc}},
\end{equation}
where $T_{\text{base}}$ is a baseline teacher and $T_{\text{cc}}$ is a constraint-coupled teacher.

Matched students can then be distilled from both teachers under the same budget and evaluation conditions. The relevant comparisons are not only task accuracy, but also robustness under longer reasoning trajectories, response stability under perturbation, ease of adversarial retuning, and hidden-burden proxies where observable.

This framework yields several public hypotheses.

\paragraph{Hypothesis 1.}
Students distilled from $T_{\text{cc}}$ should exhibit a sharper trade-off between capability retention and stability omission than students distilled from $T_{\text{base}}$.

\paragraph{Hypothesis 2.}
When path dependence is materially important for teacher competence, students trained only on teacher outputs should preserve benchmark performance less reliably under longer-horizon or perturbed evaluation settings.

\paragraph{Hypothesis 3.}
Constraint-coupled teachers should be harder to functionally compress into equally capable but weakly governed students than baseline teachers matched for nominal task accuracy.

\paragraph{Hypothesis 4.}
If a student does recover teacher-level capability from a constraint-coupled teacher, then it should also recover more of the teacher's stability profile than in an unconstrained baseline setting.

These hypotheses are falsifiable. Negative results would weaken the paper's central claim. Positive results would strengthen the case for architecture-level distillation resistance.

\section{Relation to Alignment, Watermarking, and Access Control}

This framework is complementary to existing defenses.

Alignment methods improve the desirability of model behavior \cite{christiano2017,ouyang2022,bai2022}. Access control raises the cost of extraction. Watermarking and provenance techniques help with attribution and deterrence \cite{kirchenbauer2023}. All of these remain useful.

What they generally do not guarantee is that useful capability cannot be separated from its stabilizing structure. A well-aligned model may still expose compressible behavior. A watermarked model may still transfer useful competence. A tightly monitored API may still leak enough signal for partial imitation.

Constraint coupling addresses a different layer. It asks whether the most valuable capability in a model family can be made more dependent on the latent route by which the model reasons, so that shortcut transfer becomes less effective when that route structure is not preserved.

This does not make alignment or governance obsolete. On the contrary, architecture-level resistance should be seen as one component within a larger governance stack.

\section{Trade-Secret-Safe Disclosure Boundary}

This arXiv version is intentionally public-safe. The following classes of detail are omitted:

\begin{itemize}
    \item proprietary internal variable definitions,
    \item confidential burden metrics and threshold schedules,
    \item hidden-state instrumentation procedures,
    \item implementation-specific feasibility operators,
    \item private training curricula and optimization schedules,
    \item deployment architecture and enforcement logic,
    \item system prompts, operational policies, or commercial controls,
    \item empirical results not yet cleared for public release.
\end{itemize}

These omissions are not a weakness in the paper's logic. They define its disclosure boundary. The purpose of a public version is to state the theoretical mechanism clearly enough for scientific discussion, without releasing sensitive technical details that would collapse the distinction between a theory paper and an implementation blueprint.

In that sense, the contribution is similar to a public cryptographic or safety architecture note that specifies the defensive principle while withholding operational secrets.

\section{Limitations}

Several limitations should be stated directly.

First, this is a theory paper. It does not present completed empirical validation at frontier scale.

Second, the framework depends on whether concrete burden and stability proxies can be defined in a way that is both measurable and scientifically meaningful.

Third, no claim is made that all forms of model extraction become impossible. The proposed benefit is a change in the compression frontier, not universal immunity.

Fourth, constraint coupling may introduce performance, training, or deployment costs. A future implementation must show that the safety gain justifies those costs.

Fifth, architecture cannot replace institutional governance. Evaluation, monitoring, access policy, and legal protection remain necessary even if the architectural thesis succeeds.

These limitations do not negate the framework. They define its proper scientific scope.

\section{Discussion}

The broader importance of this work lies in how it reorders AI safety design. Much current practice assumes a capable model core first and governance wrappers afterward. That order is understandable because it is modular and commercially practical. It also preserves a structural vulnerability: once a highly capable core exists, other actors can attempt to recover parts of it under cheaper or weaker control regimes.

The alternative proposed here is to make stability part of competence formation itself. If useful behavior arises through constrained latent routes rather than merely from a flexible output surface, then the model's most valuable capability becomes less cleanly separable from its internal discipline. Distillation still may occur, but the transfer problem becomes harder in precisely the dimension that matters for governance.

This is why the public version adopts a narrower but stronger claim. It does not claim an absolute solution. It claims that there exists a plausible architectural path by which distillation can lose part of its strategic advantage as a shortcut for capability transfer. That is a claim worth testing.

\section{Conclusion}

This paper presented a public, trade-secret-safe theory of distillation resistance based on constraint-coupled reasoning architectures. The starting point was a governance problem: useful capability is often more compressible than the stability and control structure that originally constrained it. Existing safety methods are important but do not fully solve that asymmetry.

To address this gap, the paper introduced a public mathematical scaffold built around bounded transition burden, path-load accumulation, dynamic feasible regions, and a capability--stability coupling condition. The resulting framework is intentionally high level. It omits confidential implementation details while preserving the central architectural thesis: when competence depends on regulated latent routes, capability becomes less easily separable from stability.

The contribution is therefore not a claim of finished proof or released system design. It is a falsifiable public theory. For an arXiv audience, that is the correct form of contribution. It defines the problem precisely, states the mechanism clearly, and marks out a research program for future empirical work on distillation resistance and model governance.

\end{document}